\documentclass{article}

\usepackage[final,nonatbib]{nips_2018}
\usepackage[utf8]{inputenc} 
\usepackage[T1]{fontenc}    
\usepackage{url}            
\usepackage{booktabs}       
\usepackage{amsfonts}       
\usepackage{nicefrac}       
\usepackage{microtype}      

\usepackage{graphicx}
\usepackage{epstopdf}
\usepackage{float}
\usepackage{amsmath}
\usepackage{amssymb}
\usepackage{color,soul}
\usepackage{multirow}
\restylefloat{table}

\usepackage{enumitem}
\usepackage{verbatim}
\setlist[itemize,1]{label={\fontfamily{cmr}\fontencoding{T1}\selectfont\textbullet}}

\newcommand{\scaption}[1]{\caption{\small{#1}}}
\newcommand{\qsection}[1]{\vspace{5pt} \noindent \textbf{#1:}}

\title{Learning to Unlearn: Building Immunity to Dataset Bias in Medical Imaging Studies}

\vspace{-25mm}
\author{Ahmed Ashraf \\ University of Manitoba \\ \texttt{Ahmed.Ashraf@umanitoba.ca}  \And Shehroz Khan \\ Toronto Rehabilitation Institute \\  \texttt{Shehroz.Khan@uhn.ca} \vspace{-20mm} \AND  Nikhil Bhagwat \\ University of Toronto \\ \texttt{nikhil.bhagwat@mail.utoronto.ca} \AND Mallar Chakravarty \\ McGill University \\ \texttt{mallar.chakravarty@douglas.mcgill.ca}  \And Babak Taati \\ Toronto Rehabilitation Institute \\ \texttt{Babak.Taati@uhn.ca} }
\begin{document}

\maketitle

\begin{abstract}
\vspace{-3mm}
Medical imaging machine learning algorithms are usually evaluated on a single dataset. Although training and testing are performed on different subsets of the dataset, models built on one study show limited capability to generalize to other studies. While database bias has been recognized as a serious problem in the computer vision community, it has remained largely unnoticed  in medical imaging research. Transfer learning thus remains confined to the re-use of feature representations requiring re-training on the new dataset. As a result, machine learning models do not generalize even when trained on imaging datasets that were captured to study the same variable of interest. The ability to transfer knowledge gleaned from one study to another, without the need for re-training, if possible, would provide reassurance that the models are learning knowledge fundamental to the problem under study instead of latching onto the idiosyncracies of a dataset. In this paper, we situate the problem of dataset bias in the context of medical imaging studies. We show empirical evidence that such a problem exists in medical datasets. We then present a framework to unlearn study membership as a means to handle the problem of database bias. Our main idea is to take the data from the original feature space to an intermediate space where the data points are indistinguishable in terms of which study they come from, while maintaining the  recognition capability with respect to the variable of interest. This will promote models which learn the more general properties of the etiology under study instead of aligning to dataset-specific peculiarities. Essentially, our proposed model learns to unlearn the dataset bias.
\end{abstract}
\vspace{-7mm}
\section{Introduction}
\vspace{-3mm}
The purpose of this paper is to situate the problem of dataset bias in the context of medical imaging, and to present a framework to handle the problem. The availability of datasets has played a central role in advancing the state of the art for learning algorithms in general, and for imaging research in particular. However, all datasets are finite attempts at encapsulating the practically infinite world to allow the imaging community a playing field  for evolving and testing their algorithms. In practice, every dataset inadvertently brings along its idiosyncrasies due to study participant selection, image acquisition variabilities, and annotation biases. This has given rise to the dataset bias problem \cite{Toralba_2011,Toralba_2012,Tatiana_2013}. Perhaps the first telling introduction to the problem was by Torralba and Efros \cite{Toralba_2011}. They demonstrated the problem by learning a classifier to name the dataset of a random image from 12 object recognition datasets. The classifier was trained on randomly sampled 1000 images from each dataset, and surprisingly it performed with a classification accuracy of 39\%, which is significantly better than chance (1/12 = 8\%). They also went on to show that with more training data, the classification accuracy could be increased with no immediate signs of saturation. 
%
%

This is a serious problem and it essentially boils down to the following. Given images with the same object category, one could tell which dataset they belong to; this means that every dataset leaves a footprint on its images, and during the course of learning a classifier, we are inadvertently learning idiosyncrasies of the dataset along with learning general properties of the object category. Does a similar situation arise in the context of medical imaging studies? We will investigate this shortly. In this paper, we identify a hitherto unnoticed link between handling dataset bias and methods to preserve privacy for fair machine learning. We draw our motivation from one particular method which aims at learning fair representations by taking the input features
to a latent space such that datapoints with
different values of sensitive information (e.g., race or gender) become
as indistinguishable as possible while maintaining
discriminablity with respect to a classification task
\cite{RICH:Fair}. \textit{We propose to adapt this method such that
we learn a latent space in which the data points
are indistinguishable in terms of the dataset they
belong to, while satisfying the classification requirements}. Two main differences between our work and \cite{RICH:Fair} must be noted. First, the sensitive variable in \cite{RICH:Fair} is binary and we modify the objective function to handle any number of datasets. Second, the classification part of the objective in \cite{RICH:Fair} is also geared for binary classification, and we generalize this part as well to handle more than two classes. 
One of the earliest approaches to explicitly handle the dataset bias problem for object recognition in visual databases was presented by Khosla et al. \cite{Toralba_2012}. In this work, the authors learn a support vector machine (SVM) classifier using images from multiple datasets and decompose the SVM weights into two parts: one part is dataset specific, while the other part is common to all datasets. A notable work on domain adaptation for medical imaging is \cite{WachingerR16}. The work that comes closest in spirit to the proposed method in this paper is domain adversarial training of neural networks \cite{DANN}. However, the methods in \cite{WachingerR16} and \cite{DANN} both use a part of target dataset for training. \textit{The method presented in this paper does not use the target dataset at all; our experiments are purely leave-one-dataset-out.} Through this paper we hope to introduce the link between dataset bias and fair machine learning, as well as encourage more work along this line to better address the problem of dataset bias.
%
\vspace{-5mm}
\section{Unlearning Dataset Membership}
\label{Sec:Model}
\vspace{-6mm}
\qsection{Notation} $\mathbf{X}$ is a  $M \times N$ data matrix, where $M$ is the number of examples, and $N$ is the number of features. $\mathbf{x}_m$ is the $m$-th data point such that $\mathbf{x}_m \in \mathbb{R}^N$. $D$ is the number of datasets. $\mathbf{X}$ contains data from different datasets. Let $s_m$ represent which dataset $\mathbf{x}_m$ belongs to. $M_d$ is the number of data points in dataset $d$. $y_m$ is the class label of the $m$-th data point. Let there be $C$ classes. The objective is to train a machine learning model which can correctly classify unseen data points, i.e., estimate a label $y$ given a new $\mathbf{x}$. $Z$ is a multinomial random variable that can assume a value $k \in [1,K]$, where a particular value $k$ represents one of the intermediate set of prototypes. Associated with each prototype is a vector $\mathbf{v}_k \in \mathbb{R}^N$.

%
%
%
\vspace{-3mm}
\qsection{Bias Unlearning Objective} Our goal is to map $\mathbf{X}$ onto a latent space, in which dataset membership is obfuscated, while classification ability is retained. Mapping from the original feature space to the latent space is via assignment to embeddings placed in the feature space. These embeddings will be referred to as ``prototypes'' in accordance with the terminology in \cite{RICH:Fair}. The objective is to learn the prototypes $\mathbf{v}_k$ such that the knowledge that a given data point gets mapped to a particular prototype does not give away any information about which dataset the data point comes from. Thus, by mapping the data points to the prototypes, we would be unlearning dataset membership. In order to formally model the above objective, we treat the prototype mapping to be soft (similar to \cite{RICH:Fair}), i.e., the mapping would be modeled as probabilities. Let $P(s =d | Z = k )$ be the probability that any arbitrary $\mathbf{x}$ belongs to dataset $d$ given that $\mathbf{x}$ was mapped to the $k$-th prototype $\mathbf{v}_k$ (i.e., $Z=k$). Now the questions is, what should be the form of the above probability distribution to satisfy the objective of not giving away any information about dataset membership with the knowledge that $Z=k$. If this distribution is uniform, our job is done; knowing $Z=k$ tells us nothing about which dataset $\mathbf{x}$ comes from. We should note that for a discrete probability distribution, the entropy is maximized when the distribution is uniform. In other words, if we select our prototypes $\mathbf{v}_k$, such that the entropy of the distribution $P(s = d | Z = k )$ (over all $k$) is maximized, our objective is satisfied. We can write this objective as: $J = - \sum_{k=1}^K  \sum_{d=1}^D P(s = d | Z = k ) \; logP(s=d | Z = k )$. To derive an expression for $P(s=d | Z = k )$, consider the following. Given the prototypes $\mathbf{v}_k$ for $k=\{1,\cdots,K\}$ and a feature vector $\mathbf{x}_m$, the probability $\mathbf{x}_m$ gets mapped to the prototype $k$ (i.e., $Z=k$) can be modeled by a normalized exponential or softmax as $P(Z = k | \mathbf{x}_m) = {e^{-\parallel \mathbf{x}_m - \mathbf{v}_k\parallel^2}} \slash { \sum_{j=1}^K e^{-\parallel \mathbf{x}_m - \mathbf{v}_j\parallel^2}}$. Other choices for distance between $\mathbf{x}_m$ and $\mathbf{v}_k$ are possible, but in this paper we stick to an unweighted Euclidean distance for simplicity. Let $\phi_{dk}$ be the expected value of the above probability over dataset $d$, i.e., $\phi_{dk} = \underset{d=D}{\mathbb{E}} P(Z = k | \mathbf{x}_m) 
 =  \frac{1}{M_d} \sum_{m=1}^M P(Z = k | \mathbf{x}_m)I(s_m=d)$; where $I(s_m=d)$ is an indicator function which equals one when $s_m=d$, and zero otherwise, to ensure that the summation is only over examples of dataset $d$. To get the overall probability of $Z=k$ for dataset $d$ i.e. $P(Z=k | s = d )$, we need to normalize $\phi_{dk}$ over all datasets. With a straight forward application of Bayes' theorem, it can now be shown that, $P( s = d | Z = k ) = \frac{ \phi_{dk} P( s = d)}{\sum_{r=1}^D \phi_{rk} P( s = r)}$. This expression can be plugged in the equation for $J$,  to get the bias unlearning objective which, if maximized with respect to the choice of prototypes $\mathbf{v}_k$, will unlearn dataset membership. 

\vspace{-2mm}
\qsection{Reconstruction constraint} We should note that our original purpose to unlearn dataset membership was to remove dataset bias. Is there a trivial solution that serves the purpose? If all data points are mapped to the same prototype, we have a trivial solution in the form of all prototypes $\mathbf{v}_k$ taking on the same position far away from all the data points such that practically the mapping does not give any information about dataset membership. A possible way to avoid such trivial solutions is to learn prototypes which allow us to (approximately) reconstruct the data. This will allow to lose just enough information to unlearn dataset membership, but will avoid extreme and trivial solutions. Considering the mapping probabilities, $P(Z=k|\mathbf{x}_m)$, as a representation of the data-point in a $K$ dimensional space, we can strive for prototypes which encourage data reconstruction. First, for brevity of notation, let us use $\psi_{mk}$ to denote $P( Z = k | \mathbf{x}_m)$. Let us define $\hat{\mathbf{x}}_m$, the reconstructed version of $\mathbf{x}_m$, as: $\hat{\mathbf{x}}_m = \sum_{k=1}^K \psi_{mk} \mathbf{v}_k$. The average reconstruction error over all the examples can now be written as: $E = \dfrac{1}{M}\sum_{m=1}^M ( \mathbf{x}_m - \hat{\mathbf{x}}_m)^2$. 
In parallel to maximizing $J$, if we learn prototypes $\mathbf{v}_k$ that minimize $E$ as above, not only would we be avoiding trivial solutions, but we would be forcing a solution such that the probabilities $\psi_{mk}$ can be treated as feature representation of our data points in a latent space.
%
%

\vspace{-2mm}
\qsection{Classification loss} While we want to unlearn dataset membership, at the same time we intend to maintain classification capability. The reconstruction constraint in the previous section allows interpreting the probabilities $\psi_{mk}$ as features in the intermediate space. In order to enable classification, we now add multiclass cross-entropy loss, using $\psi_{mk}$ as features. Let the $K$ dimensional representation of the $m$-th data point be represented as $\mathbf{\Psi}_m = [ \psi_{m1} \;\; \psi_{m2} \;\; \cdots \; \psi_{mK} ]^T$.
Let the softmax weights for the $c$-th class ($c=\{1, \cdots C\}$) be $\mathbf{\Theta}_c=[ \theta_{c1} \;\; \theta_{c2} \;\; \cdots \; \theta_{cK} ]^T$. The average classification loss can then be written as, $L = \dfrac{1}{M} \sum_{m=1}^M \sum_{c=1}^C I(y_m = c) \; log \dfrac{ \text{exp}( \mathbf{\Theta}^{T}_c \mathbf{\Psi}_m)}{ \sum_{t=1}^C \text{exp}( \mathbf{\Theta}^{T}_t \mathbf{\Psi}_m)}$. We should note that the classification loss is not only a function of the softmax weights, but also of the prototypes $\mathbf{v}_k$, as the feature representation $\mathbf{\Psi}_m$ depends on them.
\vspace{-2mm}

%
%
%
%
%
%

\qsection{Combined Loss} The overall objective can be met if we minimize the following loss function with respect to $\mathbf{v}_k$ and $\mathbf{\Theta}_c$: $\mathcal{L} = -\alpha_J J + \alpha_E E + \alpha_L L$; where $J$, $E$, and $L$ are the bias unlearning objective, reconstruction error, and classification loss respectively. Here $\alpha_J$, $\alpha_E$, and $\alpha_L$ are hyperparameters indicating the relative weight of each term, and could be determined using cross-validation.

\vspace{-4.5mm}
\section{Datasets}
\vspace{-4mm}
To test our proposed method, we selected four brain MRI datasets which are actively employed to benchmark algorithms for detecting imaging correlates of cognitive health. Three datasets come from the three phases of the The Alzheimer's Disease Neuroimaging Initiative (ADNI) \cite{ADNI_2010}. These datasets are referred to as ADNI 1, ADNI 2, and ADNI GO. The fourth dataset comes from the Australian Imaging, Biomarker and Lifestyle Flagship Study of Aging (AIBL) \cite{AIBL_2009}. For ADNI studies, a subset of participants was selected such that there was no overlap of participants amongst datasets. The imaging feature space consisted of 78 features, as derived from 78 regions of interest (ROIs) based on Automatic Anatomical Labeling (AAL) atlas \cite{SPM_2002}. For more details please refer to \cite{ADNI_2010}, \cite{AIBL_2009}.
\vspace{-5mm}
\section{Experiments}
\label{Sec:Experiments}
\vspace{-5mm}
\qsection{``Name the study'' classifier} We first attempt to build a four class classifier to test if dataset bias has crept even into the MRI datasets based on AAL atlas features. Specifically, we train a four class softmax (one of the simplest models) using a two-fold cross validation. The overall classification accuracy is 65.92\% which is considerably higher than chance performance of 25\%. Only examples from ADNI-GO dataset presented difficulty in identifying their dataset membership.

\vspace{-3mm}
\qsection{Mild Cognitive Impairment (MCI) versus Healthy Classification} As an output variable of interest, in this paper we focus on the task of distinguishing MCI patients from cognitively healthy patients. To evaluate cross-dataset generalization, we used a leave-one-dataset-out (LODO) strategy, both for learning the prototypes and classifiers. As such, within a particular LODO experiment, learning was done on the basis of three datasets, while testing was done on the left out dataset. As a performance metric, we used the area under the curve (AUC) for the classifier receiver operating characteristics (ROC). Since ADNI-GO consisted of examples from only one class, AUC could not be computed when it was the left out dataset. However, ADNI-GO was still used for training when the left out dataset was one of ADNI1, ADNI2, or AIBL. Our approach has four hyperparameters: $\alpha_J$, $\alpha_E$, $\alpha_L$, and the $L_2$ regularization penalty $\lambda$ for the softmax classification in the latent space. For hyperparameter tuning we performed a grid search by running a nested LODO amongst the training datasets, and optimizing for the ROC-AUC. 
%
%
%

\vspace{-3mm}
\qsection{Baseline 1 $-$ Na\"{i}ve LODO} As a bare minimum, a cross-dataset generalization algorithm should outperform the baseline of a na\"{i}ve LODO strategy. For this purpose we trained a logistic regression MCI versus healthy classifier with LODO over the four datasets under consideration.

\vspace{-3mm}
\qsection{Baseline 2 $-$ Unbiased SVM \cite{Toralba_2012}}
Khosla et al. \cite{Toralba_2012} have shown superior performance to na\"{i}ve leave-one-dataset-out approach for visual recognition tasks.  The authors modify the SVM objective function such that two kinds of weights are learned: dataset specific, and weights common to all datasets. The common weights are supposed to learn the more general structure of the problem and are used to classify examples from the left out dataset. We trained the unbiased SVM with LODO for MCI versus healthy classification.

%
%
%
%
\vspace{-3mm}
\qsection{Performance ``Ceiling''} While not necessary, it is reasonable to expect that within-dataset classification, wherein the test and training examples come from the same dataset, would perform better than a cross-dataset classifier. To examine this, we trained traditional within-dataset MCI vs Healthy classifiers for each dataset using a two-fold cross validation. More specifically, to measure the goodness of a cross-dataset method we compute the drop in performance while doing LODO classification as a percentage of within-dataset classifier performance. 

\vspace{-5.5mm}
\section{Results}
\label{Sec:Results}
\begin{figure}[h]
\centering
\includegraphics[scale=0.45]{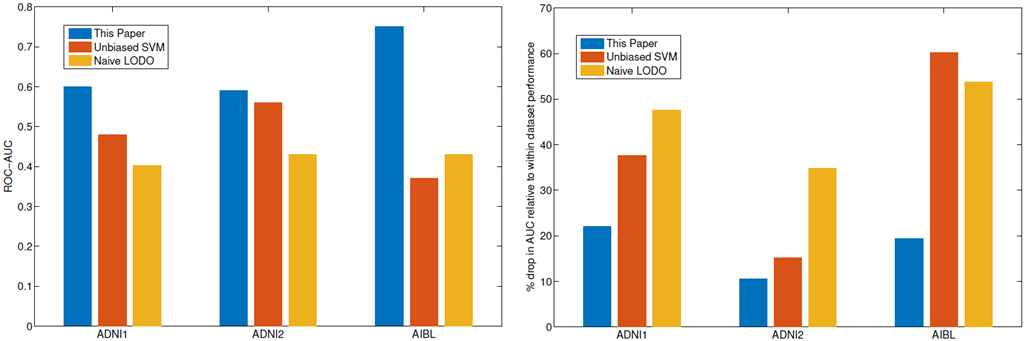}
\vspace{-3mm}
\scaption{Comparison of different leave-one-dataset-out approaches in terms of ROC-AUC (left), and drop in ROC-AUC as a percentage of within-dataset-classification performance (right)}
\label{Fig:BAR1}
\end{figure}

\vspace{-4mm}
Figure \ref{Fig:BAR1}(left) shows the comparison of different methods in terms of ROC-AUC for MCI versus healthy classification. For all datasets, unlearning dataset membership gives rise to better cross-dataset generalization. The traditional within-dataset classification approach performs with AUCs of 0.77, 0.66, and 0.93 for ADNI1, ADNI2, and AIBL datasets respectively. These numbers, not surprisingly, are higher than those for LODO strategies. We should note that na\"{i}ve LODO performs worse than chance for all datasets; unbiased SVM works above chance for only one dataset. Our method always performs better than chance. A comparison in terms of drop in AUC as a percentage of within-dataset performance is given in Figure \ref{Fig:BAR1}(right). By unlearning dataset membership the drop remains less than 20\%, while for other approaches the drop is higher. 
%
%
%
%
%
%
%
%
%

\vspace{-3.9mm}

\vspace{-8.5mm}
\section{Discussion}
\vspace{-5mm}
In this paper we have presented a method to unlearn dataset membership for handling dataset bias while attempting to maintain the classification ability. We have presented leave-one-dataset out results using four brain MRI datasets to study cognitive health. Our results show that AUC values improve on average by 47 and 18 points in comparison to the baselines of simple leave-one-dataset-out and unbiased SVM respectively. Moreover, by unlearning dataset membership the drop in performance as compared to within-dataset classification is also minimum. ADNI1 and ADNI2 datasets were easily identifiable in the ``name the study'' experiments, i.e. they have more idiosyncracies, and cross-dataset generalization on ADNI1 and ADNI2 is less than that of AIBL which is more similar to other datasets (got confused with ADNI1 and ADNI2 in dataset identification). If two datasets are similar, the results from one dataset would generalize well to the other dataset. Our results are in line with this intuition. 
%
%
%
%
%
%
%
%


\vspace{-5mm}
\bibliographystyle{plain}
\bibliography{ML4H}
\end{document}